\documentclass[conference]{IEEEtran}
\IEEEoverridecommandlockouts
\usepackage{cite}
\usepackage{amsmath,amssymb,amsfonts}
\usepackage{algorithmic}
\usepackage{graphicx}
\usepackage{textcomp}
\usepackage{xcolor}
\usepackage{bm}
\usepackage{multirow}
\usepackage{colortbl}
\usepackage{booktabs}

\usepackage{pgfplots}
\usetikzlibrary{patterns,backgrounds}
\usepgfplotslibrary{groupplots}

\def\BibTeX{{\rm B\kern-.05em{\sc i\kern-.025em b}\kern-.08em
    T\kern-.1667em\lower.7ex\hbox{E}\kern-.125emX}}
\begin{document}

\title{An End-to-End Model for Photo-Sharing Multi-Modal Dialogue Generation}

\author{
\thanks{This work was supported by the National Natural Science Foundation of China under Grant 62336008 and the Shenzhen Science and Technology Program under Grant GXWD20231130140414001.}

\IEEEauthorblockN{Peiming Guo\IEEEauthorrefmark{1}, Sinuo Liu\IEEEauthorrefmark{2}, Yanzhao Zhang\IEEEauthorrefmark{3}, Dingkun Long\IEEEauthorrefmark{3}, Pengju Xie\IEEEauthorrefmark{3}, Meishan Zhang\IEEEauthorrefmark{1}\IEEEauthorrefmark{4}\thanks{\IEEEauthorrefmark{4} Corresponding author}, Min Zhang\IEEEauthorrefmark{1}}
\IEEEauthorblockA{\IEEEauthorrefmark{1}Harbin Institute of Technology (Shenzhen), \IEEEauthorrefmark{2}The University of Edinburgh, \IEEEauthorrefmark{3}Independent Researcher}
\IEEEauthorblockA{guopeiming.gpm@gmail.com, sinuo.liu@outlook.com, zhangyanzhao00@gmail.com, longdingkun1993@gmail.com, \\ xpjandy@gmail.com, mason.zms@gmail.com, zhangmin2021@hit.edu.cn}

}

\maketitle

\begin{abstract}
Photo-Sharing Multi-modal dialogue generation requires a dialogue agent not only to generate text responses but also to share photos at the proper moment.
Using image text caption as the bridge, a pipeline model integrates an image caption model, a text generation model, and an image generation model to handle this complex multi-modal task.
However, representing the images with text captions may lose important visual details and information and cause error propagation in the complex dialogue system.
Besides, the pipeline model isolates the three models separately because discrete image text captions hinder end-to-end gradient propagation.
We propose the first end-to-end model for photo-sharing multi-modal dialogue generation, which integrates an image perceptron and an image generator with a large language model.
The large language model employs the vision encoder to perceive visual images in the input end.
For image generation in the output end, we propose a dynamic vocabulary transformation matrix and use straight-through and gumbel-softmax techniques to align the large language model and stable diffusion model and achieve end-to-end gradient propagation.
We perform experiments on PhotoChat and DialogCC datasets to evaluate our end-to-end model.
Compared with pipeline models, the end-to-end model gains state-of-the-art performances on various metrics of text and image generation.
\end{abstract}


\section{Introduction}

\begin{figure}[t]
    \centering
    \includegraphics[width=0.5\textwidth]{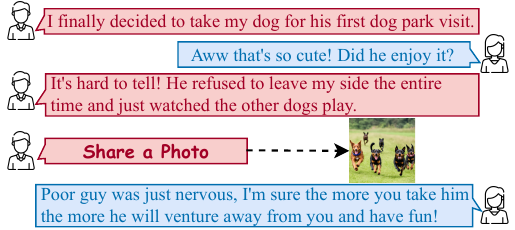}
    \caption{Photo-sharing multi-modal dialogue.}
\label{fig:intro}
\end{figure}

Human-machine dialogue system is a crucial research and application area for the artificial intelligence community. 
With the development of vision-language representation learning, multi-modal human-machine dialogue has attracted widespread attention~\cite{Das_2017_CVPR,shuster-etal-2020-image,zheng-etal-2022-mmchat}.
However, the multi-modal settings in these literatures and recent multi-modal large language models (MLLMs) (e.g., Llama-Vision~\cite{llama3modelcard}, Qwen-VL~\cite{Qwen-VL}) are limited to adding visual modality to the input, while the output of the dialogue model remains pure textual modality.
Similar to the conversation manner of humans on instant messaging softwares (e.g., Facebook, WhatsApp, and WeChat), an engaging multi-modal dialogue agent should not only perceive and comprehend visual content but also generate images.
Therefore, researchers have started to focus on \emph{photo-sharing} multi-modal dialogue in recent years~\cite{lee-etal-2021-constructing,zang-etal-2021-photochat,sun-etal-2022-multimodal,feng-etal-2023-mmdialog,lee2023dialogcc,aboutalebi2024magid}, which requires the active behavior of sharing photos in multi-turn dialogue like \figurename~\ref{fig:intro} for multi-modal chatbots.

In order to build the photo-sharing multi-modal dialogue system, \cite{sun-etal-2022-multimodal} proposes a pipeline model, Divter, which composes three modules: image caption model, text dialogue model, and image generation model.
The pipeline model utilizes the image text description as the bridge to connect visual and textual modality.
Although the pipeline model can achieve photo-sharing multi-modal dialogue, the system still faces many challenges.
First, the text dialogue model may lose visual details in the photos, because the image caption model will significantly compress visual information during the process of generating image descriptions.
As a result, it is difficult for the model to produce elaborate utterance replies.
Second, the pipeline model suffers from the severe problem of error propagation.
Any small mistake in the image descriptions will be propagated and enlarged into the generated text dialogue responses and images.
Third, the pipeline model includes three neural network models, which are independent and cannot achieve end-to-end gradient propagation and training.
So the separate training of these models makes it difficult to capture the internal relationships between different modules.
Besides, the cascaded system has redundant functions and parameters, leading to latency and inefficiency in practical deployment.

In this work, we propose an end-to-end photo-sharing multi-modal dialogue model to alleviate the challenges in the pipeline model, resorting to recent MLLMs, diffusion models~\cite{esser2024scaling} and end-to-end gradient connection techniques (Straight-Through Estimator~\cite{bengio2013estimating} and Gumbel-Softmax~\cite{jang2017categorical,maddison2017the}).
For one thing, we use the MLLM (i.e., Llama-Vision) as the backbone of dialogue generation, which enables the realization of the end-to-end multi-modal model at the input end.
For another thing, the stable diffusion~\cite{esser2024scaling} is engaged in image generation.
Especially, inspired by~\cite{chen2024semantic} and~\cite{lai2024lisa}, we design a dynamic vocabulary transformation matrix to convert LLM vocabulary into diffusion model vocabulary, which integrates Straight-Through Estimator~\cite{bengio2013estimating} and Gumbel-Softmax~\cite{jang2017categorical,maddison2017the} to connect the diffusion model with the MLLM and realize the end-to-end gradient connection at the output end.
Overall, the MLLM and the diffusion model equip the large language model with the ability to \emph{see} and \emph{draw} images at the same time.

In conclusion, the main contributions can be summarized as follows:
\textbf{i)} We propose the first end-to-end photo-sharing multi-modal dialogue model, which jointly trains the MLLM and diffusion model to acquire the abilities of image perceiving, text generation and image generation.
\textbf{ii)} We design a novel dynamic vocabulary transformation matrix and apply straight-through and gumbel-softmax techniques to convert image text description from LLM to diffusion model while retaining gradients.
\textbf{iii)} Experimental results on diverse metrics verify the effectiveness of our end-to-end model compared with pipeline models~\footnote{We public our code at https://github.com/guopeiming/E2E\_PSDG.}.

\section{Related Work}
With the development of vision-language representation learning in this decade, multi-modal dialogue has attracted widespread attention~\cite{Das_2017_CVPR,shuster-etal-2020-image,zheng-etal-2022-mmchat}.
However, conventional multi-modal dialogue tasks consider vision modality only at the input end.
Therefore, researchers have focused on photo-sharing multi-modal dialogue in recent years~\cite{lee-etal-2021-constructing,zang-etal-2021-photochat,sun-etal-2022-multimodal,feng-etal-2023-mmdialog,lee2023dialogcc,aboutalebi2024magid}.
\cite{zang-etal-2021-photochat} proposes a retrieval model to predict the image shared in the dialogue, which computes the similarity between dialogue context and image.
\cite{feng-etal-2023-mmdialog} also models photo-sharing multi-modal dialogue by a retrieval task. 
Different from only predicting images in~\cite{zang-etal-2021-photochat}, \cite{feng-etal-2023-mmdialog} retrieves suitable textual utterances and shared photos for conversations.
In addition to the retrieval model, \cite{sun-etal-2022-multimodal} presents a generative model, Divter, for photo-sharing multi-modal dialogue generation.
Specifically, Divter adapts a text dialogue model as the backbone, and perceives photos based on discrete image descriptions from an image caption model for the input end.
For the output end, the text dialogue model generates text responses and image descriptions, which are converted into photos by an image generation model.

Another important related work for this paper is multi-modal generation models, including image generation models and MLLMs.
Image generation models have gained great attention in the deep learning community in this decade.
In the beginning, researchers employ VAE and GAN models to generate image examples.
Following, VQ-VAE~\cite{10.5555/3295222.3295378} and VQ-GAN~\cite{Esser_2021_CVPR} quantize latent vectors into discrete codebook to improve VAE and GAN models.
Recently, diffusion-based models have shown better image generation ability~\cite{zhao2024synergistic}, like DALL-E-2~\cite{ramesh2022hierarchical} and stable diffusion~\cite{esser2024scaling}.
Most MLLMs~\cite{liu2023visual,Qwen-VL} integrate a vision encoder into LLMs, which learn a mapping between text modality (large language model) and vision modality (vision encoder).
However, these MLLMs acquire multi-modal capability only at the input end, employing linear projection~\cite{liu2023visual}, Q-Former~\cite{Qwen-VL} or cross-attention as~\cite{llama3modelcard} as the modality mapping adapter.
A few MLLMs~\cite{Chameleon_Team_Chameleon_Mixed-Modal_Early-Fusion_2024,ge2024making} have recently implemented visual ability at both the input and output end by vision tokenizer.
\section{Model}

\begin{figure*}
    \centering
    \includegraphics[width=\textwidth]{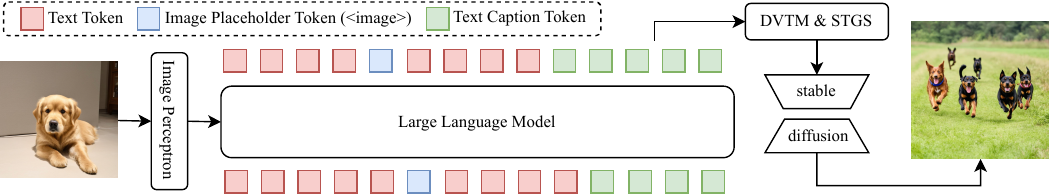}
    \caption{The overview of our end-to-end photo-sharing multi-modal dialogue generation model, which integrates an image perceptron and an image generator with a large language model. The large language model employs the image perceptron to perceive visual images in the input end. For image generation in the output end, we propose a dynamic vocabulary transformation matrix (DVTM) and use straight-through and gumbel-softmax (STGS) techniques to align the large language model and stable diffusion model and achieve end-to-end gradient propagation. }
\label{fig:model}
\end{figure*}

In this section, we start with the task formalization of photo-sharing multi-modal dialogue (\S\ref{text:task_formalization}), then introduce our end-to-end model shown in \figurename~\ref{fig:model}, which consists of three components: text dialogue generator (\S\ref{text:text_dialogue_generator}), image perception (\S\ref{text:image_perceptron}) and image generator (\S\ref{text:image_generator}), end with the description of end-to-end training (\S\ref{text:end-to-end_training}).

\subsection{Task Formalization}
\label{text:task_formalization}

Given a photo-sharing multi-modal dialogue dataset $\mathcal{D} = \{ ( U_{i}, R_{i} ) \}_{i}^{I}$, where $\forall ~i \in \{1, 2, \dots, I\}$, $U_{i}$ is the dialogue context, $R_{i}$ is the response regarding to $U_{i}$.
Concretely, for each dialogue context and response, $U_{i} = \{ u_{k} \}_{k}^{K}$ and $R_{i} = \{ r_{l} \}_{l}^{L}$ are sequences of multi-modal elements including textual utterances and visual images. 
$K$ and $L$ are the number of elements in $U_{i}$ and $R_{i}$ respectively.
Our goal is to learn an end-to-end photo-sharing multi-modal dialogue model $P(R|U)$ supervised by the dataset $\mathcal{D}$.
When given the multi-modal dialogue context $U$, the end-to-end model can generate multi-modal response $R$ including texts and images.

\subsection{Text Dialogue Generator}
\label{text:text_dialogue_generator}
MLLMs have demonstrated powerful cross-modal comprehension and text generation ability in recent years, integrating LLM and extra vision modules.
We adopt a popular MLLM, Llama-Vision~\cite{llama3modelcard}, for our end-to-end photo-sharing multi-modal dialogue model, where the pure textual LLM (i.e., Llama) is the backbone of text dialogue generation.
Suppose that $U = \{ u_{1}, \dots, u_{K} \}$ is a dialogue context, the text context in it will be tokenized and converted into vector representations by the text dialogue generation, while the image context will be encoded by the image perceptron that is introduced in the next subsection.
Similarly, we formulate the dialogue response $R = \{ r_{1}, \dots, r_{L}\}$ as the target sequence for the text dialogue generator. 
If the response element $r_{l}$ is text modality, it will be represented as a textual token sequence of the text utterance $r_{l} = \{r_{l}^{1}, \dots, r_{l}^{n}\}$.
If the response element $r_{l}$ is vision modality, it will be represented as a textual token sequence $r_{l} = \{\text{[IMG]}, r_{l}^{1}, \dots, r_{l}^{n}, \text{[/IMG]}\}$ of the image text description surrounded by special tokens $\text{[IMG]}$ and $\text{[/IMG]}$.

We input the dialogue context $U$ and response $R$ to the text dialogue generator as the source and target sequence for teacher-forcing training on the vocabulary probability distribution $\bm{p}_{i}$ of the $i$-th time step:
\begin{equation}
\begin{aligned}
    \bm{p}_{i} &= \text{LLM}(U, r_{\textless i})\\
    \mathcal{L}^{\text{t}} &= -\text{log} ~P(R|U) = -\Sigma_{i=1}^{|R|} \text{log} ~\bm{p}_{i}^{r_{i}},
\end{aligned}
\label{eq:llm}
\end{equation}
where $\bm{p}_{i}^{r_{i}}$ is probability of the token $r_{i}$.
For the inference stage, LLM auto-regressively generates textual utterances or image captions.
Particularly, the image caption and the corresponding vocabulary probability distribution $\bm{p}_{i}$ will be inputted to the image generator, which employs the diffusion model and our novel gradient connection technique to generate images and maintain gradient propagation.

\subsection{Image Perceptron}
\label{text:image_perceptron}
For the photo in the input end, the pipeline model bridges text and vision modality by discrete image text description, which interrupts gradient computation and overlooks visual details.
To enable LLMs to perceive the visual modality information of the entire image directly, the image perceptron attempts to connect a visual encoder to the text dialogue generator.
Concretely, Llama-Vision first acquires image representations by a pre-trained visual encoder and learnable projector, then inserts an image placeholder token ($\langle$image$\rangle$) into text, and finally applies cross-attention to \emph{see} the input photo.
Since the visual vectorized representations are fed to LLM directly, the image perceptron is an end-to-end model.

\subsection{Image Generator}
\label{text:image_generator}

Photo-sharing multi-modal dialogue requires the multi-modal chatbot to share a photo at the right moment actively, so we design the image generator to be responsible for the image generation function.
We adapt the stable diffusion model~\cite{esser2024scaling} in our end-to-end system for its impressive generation performance and computational efficiency.
Similar to the handling of the visual modality in the input end, the pipeline model still drives the vision model through the discrete image text description for image generation in the output end.
Our end-to-end model also employs image text description to bridge the large language model and the stable diffusion model, but we propose the Straight-Through Gumbel-Softmax~\cite{bengio2013estimating,jang2017categorical,maddison2017the} and dynamic vocabulary transformation matrix to implement end-to-end gradient propagation from the image generator to the text dialogue generator.

\subsubsection{Straight-Through Gumbel-Softmax}
\label{text:gumbel_softmax}
The pipeline model completely isolates the text dialogue generation model and the image generation model and only bridges them with the discrete image text caption. 
Based on the straight-through gumbel-softmax~\cite{bengio2013estimating,jang2017categorical,maddison2017the}, we propose an end-to-end model to connect the LLM and stable diffusion model in continuous vector space not discrete textual space.
Concretely, the text dialogue generator applies argmax operation to generate an image caption, which is then delivered to the stable diffusion model to guide photo generation.
The argmax and textual input are the sources of weak robustness and the isolation between text dialogue generation and image generation.
Compared with the argmax operation, gumbel-softmax introduces randomness into the prediction, resulting in a sampled vector.
More importantly, gumbel-softmax adapts a reparameterization trick to make the sampling process differentiable.
Specifically, we first sample random variable $\bm{g}$ and add it into the vocabulary probability distribution $\bm{p}_{i}$ to gain a new one $\bm{p}_{i}^{\text{GS}}$.
Then we apply hard gumbel-softmax to avoid the soft categorical probability, which achieves the exact one-hot output $\bm{r}_{i}$ by argmax and retains gradient by straight-through gradient estimator~\cite{bengio2013estimating}:
\begin{equation}
\begin{aligned}
    \bm{g} & \sim Gumbel(0, 1) \\
    \bm{p}_{i,j}^{\text{GS}} &= \frac{\text{exp}((\text{log}(\bm{p}_{i,j}) + \bm{g}_{i,j}) / \tau)}{\sum_{k=0}^{|V^{\text{LLM}}|} \text{exp}(\text{log}(\bm{p}_{i,k}) + \bm{g}_{i,k}) / \tau } \\
    \bm{r}_{i} &= \text{argmax}(\bm{p}_{i}^{\text{GS}}) - \text{sg}[\bm{p}_{i}^{\text{GS}}] + \bm{p}_{i}^{\text{GS}}.
\end{aligned}
\label{eq:gumbel_softmax}
\end{equation}
where $\bm{p}_{i,j}$, $\bm{p}_{i,j}^{\text{GS}}$ and $\bm{g}_{i,j}$ are the $j$-th values of the vectors $\bm{p}_{i}$, $\bm{p}_{i}^{\text{GS}}$ and $\bm{g}_{i}$, respectively. $|V^{\text{LLM}}|$ indicates the size of the LLM vocabulary $V^{\text{LLM}}$.
As the temperature $\tau$ approaches 0, samples from the gumbel-softmax distribution become one-hot.
$\text{sg}[\cdot]$ is the stop-gradient operation for straight-through gradient estimator, which not only ensures exactly discrete vector outputs but also copies gradient from $\bm{r}_{i}$ to $\bm{p}_{i}^{\text{GS}}$.
Finally, we aggregate one-hot token predictions from the large language model into an image text caption sequence $\bm{R}^{\text{LLM}} = (\bm{r}_{1}, \cdots, \bm{r}_{m})$.

\subsubsection{Dynamic Vocabulary Transformation Matrix}
\label{text:dynamic_vocab}

\paragraph{Vocabulary Transformation Matrix}
Aiming to achieve the end-to-end model, we can not convert the sampled one-hot output $\bm{R}^{\text{LLM}}$ into textual caption by LLM tokenizer and vocabulary, otherwise the gradient propagation will be interrupted.
Therefore we need to feed the vectorized output into the stable diffusion model directly.
However, the vocabularies of the LLM $V^{\text{LLM}}$ and the stable diffusion model $V^{\text{SD}}$ are mismatched.
To align the mismatched vocabularies, we propose vocabulary transformation matrix $\bm{M}$ to implement the conversion of one-hot character representations, which is defined as follows:
\begin{equation}
    \bm{M}_{i,j} =
    \begin{cases}
     1, ~~~~~~~\text{if}~~ V^{\text{LLM}}[i] = V^{\text{SD}}[j],  \\
     0, ~~~~~~~\text{otherwise,}
     \end{cases}
\end{equation}
where $V^{\text{LLM}}[i]$ and $V^{\text{SD}}[j]$ indicate the $i$-th and $j$-th token in the vocabulary of LLM and stable diffusion model, respectively.
Based on $\bm{M}$, we formalize the mapping of one-hot image text caption representation from $V^{\text{LLM}}$ to $V^{\text{SD}}$ as matrix multiplication (i.e., $\bm{R}^{\text{SD}} = \bm{R}^{\text{LLM}} \bm{M}$), which can ensure gradient propagation between different vocabularies.

\paragraph{Dynamic Vocabulary Transformation Matrix}
The above vocabulary transformation method is under a strong assumption: the corresponding relation of tokens in the two vocabularies is one-to-one.
However, the real situation is difficult to satisfy such hypothesis, because current vocabularies usually use word piece or sentence piece tokenization.
These tokenization methods split words into different tokens according to hyperparameters, corpora, tokenization algorithms, etc.
Therefore, the differences between tokens from different vocabularies are large.
Many tokens appear in only one vocabulary (i.e., one-to-zero), resulting in a very sparse matrix $\bm{M}$.
A direct method for this issue is to apply word-level corresponding relations to construct the matrix.
Concretely, we assign the mapping relation between each token in the token set of a word and all tokens in the token set of the same word, which is many-to-many relation in fact.
However, if we construct the vocabulary transformation matrix on all English words before model training, $\bm{M}$ will be very dense and raise over-mapping and mis-mapping problems.
In particular, most different tokens from the two vocabularies have mapping relations, which could not maintain exact and delicate token alignments.

To address this issue, we propose the dynamic vocabulary transformation matrix.
We dynamically construct the vocabulary matrix $\bm{M}$ during each batch computing based on the word in the sentence, not before model training based on all English words.
Constructing the vocabulary transformation matrix on only one word reduces its density and precisely ensures the mapping relationships of all tokens in the word.
However, this operation is difficult for training, and computationally unfeasible, because the vocabulary transformation matrix initialization is performed repeatedly, wasting GPU computational time and memory.
Therefore, we design the sentence-level dynamic vocabulary transformation matrix to balance dialogue performance and computational efficiency.
Specifically, given the token set of an image text caption on the large language model vocabulary and stable diffusion model vocabulary, $\mathcal{T}^{\text{LLM}}$ and $\mathcal{T}^{\text{SD}}$, we assign token alignments to all tokens of the two sets in the dynamic vocabulary transformation matrix:
\begin{equation}
    \bm{M}_{i,j} =
    \begin{cases}
     1, ~~\text{if}~~ V^{\text{LLM}}[i] \in \mathcal{T}^{\text{LLM}} \bigwedge V^{\text{SD}}[j] \in \mathcal{T}^{\text{SD}},  \\
     0, ~~\text{otherwise,}
     \end{cases}
\end{equation}
During the training phase, we initialize the vocabulary transformation matrix for each image caption in the batch dialogue.
It is important to note that the vocabulary transformation matrix can lead to significant GPU memory overhead.
Suppose the sizes of the LLM and stable diffusion model vocabularies are both 40,000, each matrix will occupy about $40,000 * 40,000 * 2 = 1,600,000,000 * 2= 3.2G$ bytes GPU space.
However, the transformation matrix is sparse factually.
We hence implement it through the special sparse tensor class in numpy and pytorch, which constrains GPU memory occupation to a magnitude of kilobytes (KB).

Although we acquire the vector representation of the image caption, it is not a standard one-hot vector format.
All positions of image caption tokens in each vector are 1, not only one position.
In addition, another intractable problem is that the length of the converted image caption is not adaptive to the tokenization of the stable diffusion model.
To address these issues, we first use the textual image caption to get a valid one-hot vector representation $\widetilde{\bm{R}^{\text{SD}}}$ with the correct length by the stable diffusion model tokenizer.
Second, we feed the converted representation $\bm{R}^{\text{SD}}$ into average pooling to eliminate sequence length dimension.
Third, straight-through gradient estimator~\cite{bengio2013estimating} is adopted to propagate gradients again:
\begin{equation}
\begin{aligned}
    \bm{R}^{\text{SD}} &= \widetilde{\bm{R}^{\text{SD}}} - \text{sg}[\text{avg}(\bm{R}^{\text{LLM}} \bm{M})] + \text{avg}(\bm{R}^{\text{LLM}} \bm{M})
\end{aligned}
\end{equation}
where avg indicates average pooling.
Finally, we employ the stable diffusion model~\cite{esser2024scaling} for image generation based on $\bm{R}^{\text{SD}}$.
For inference, we directly feed $\widetilde{\bm{R}^{\text{SD}}}$ to the stable diffusion model because gradients is not necessary.

\subsection{End-to-End Training}
\label{text:end-to-end_training}

For the input end, MLLM connects text and vision modality, which gives the LLM the ability to \emph{see}.
For the output end, the dynamic vocabulary transformation matrix integrates Straight-Through Gumbel-Softmax to implement end-to-end gradient propagation, enabling the LLM to \emph{draw} images.
Our end-to-end model realizes fluent gradient flow throughout the entire dialogue system, which can be jointly trained:
\begin{equation}
\begin{aligned}
\mathcal{L} &= \mathcal{L}^{\text{t}} + \alpha \mathcal{L}^{\text{v}},
\end{aligned}
\end{equation}
where $\alpha$ is the weighted coefficient, and $\mathcal{L}^{\text{v}}$ is the loss of the stable diffusion model for image generation.

\section{Experiments}

\subsection{Experimental Setup}

\subsubsection{Datasets} We conduct experiments on two datasets: PhotoChat~\cite{zang-etal-2021-photochat} and DialogCC~\cite{lee2023dialogcc}.
We obtain text dialogue contents from corresponding GitHub repos.
For shared photos in the datasets, we download them according to the URLs of images.
However, some image are not accessible now due to expired URLs.
We leverage stable diffusion 3~\cite{esser2024scaling} to generate the missing images based on the image text captions in datasets.
We employ standard train, dev and test dataset splits for model training, evaluation and testing, respectively.


\subsubsection{Hyperparameters}
For the text dialogue generator and image perceptron in MLLM, we exploit Llama-Vision~\cite{llama3modelcard} as the backbone.
For the image generator, we fine-tune the stable-diffusion-3-medium~\cite{esser2024scaling}.
Besides, for hyper-parameters of model training, we use the AdamW algorithm with learning rate 5e-5, batch size 32, weight decay 0.01 and linear learning rate warmup over the first 1,000 steps to optimize parameters.
The temperature parameter $\tau$ is gradually annealed from 1 to 0.0001 in the first three epochs and remains constant in the subsequent rounds.
Weight factor $\alpha$ is 1.
We run all experiments on Nvidia V100.

\subsubsection{Evaluation}
We evaluate our end-to-end model on both text and image generation for the photo-sharing multi-modal dialogue task.
For text generation, we leverage BLEU-1, BLEU-2 and Rouge-L as evaluation metrics.
For image generation, we use Inception Score (IS) and Frechet Inception Distance (FID).
We conduct the experiments on three different random seeds and report the average results.

\subsubsection{Baselines}
To verify the effectiveness of our end-to-end model, we compare results with two related baseline models:
1) Divter \cite{sun-etal-2022-multimodal} uses DialoGPT and DALL-E to implement the pipeline-based model.
2) pipeline adapts the same Llama-3 and stable-diffusion-3-medium as the end-to-end model, but trains them separately.
3) GPT-4~\cite{achiam2023gpt} and 4) Chameleon~\cite{Chameleon_Team_Chameleon_Mixed-Modal_Early-Fusion_2024} are two representative end-to-end MLLMs. 

\subsection{Main Results}

\begin{table}[t]
    \centering
    \caption{Main results for text and image response generation. $^\dagger$ indicates the model is trained on extra resources of image generation.}
    \begin{tabular}{c|ccc|cc}
    \toprule
    \multirow{2}{*}{\textbf{Model}} & \multicolumn{3}{c|}{\textbf{Text Generation}} & \multicolumn{2}{c}{\textbf{Image Generation}} \\
    & \textbf{BL-1} & \textbf{BL-2} & \textbf{Ro-L} & \textbf{IS $\uparrow$} & \textbf{FID $\downarrow$} \\
    \midrule
    \rowcolor{gray!30} \multicolumn{6}{c}{PhotoChat} \\
    Divter~\cite{sun-etal-2022-multimodal}$^\dagger$  & 6.52 & 1.66 & 5.69 &  \textbf{15.8} $\pm$ \textbf{0.6}  & \textbf{29.16}  \\
    Divter~\cite{sun-etal-2022-multimodal} & 6.28 & 1.51 & 5.40 &  4.9 $\pm$ 0.7  & 262.09  \\
    GPT-4~\cite{achiam2023gpt} & 5.78 & 1.02 & 4.97 &  12.66 $\pm$ 1.05 & 131.40 \\
    Chameleon~\cite{Chameleon_Team_Chameleon_Mixed-Modal_Early-Fusion_2024} & 8.62 & 1.73 & 7.88 &  6.04 $\pm$ 1.63 & 177.79 \\
    pipeline & 10.68 & 2.10 & 9.74    &  13.71 $\pm$ 1.35  & 79.94 \\
    end-to-end & \textbf{12.14} & \textbf{2.98} & \textbf{11.12} &  14.59 $\pm$ 0.91  & 75.84 \\

    \rowcolor{gray!30} \multicolumn{6}{c}{DialogCC} \\
    Divter~\cite{sun-etal-2022-multimodal} & 8.08 & 1.89 & 6.71 &  9.19 $\pm$ 2.01  & 108.50 \\
    GPT-4~\cite{achiam2023gpt} & 6.13 & 0.87 & 4.04 & 11.50 $\pm$ 1.97 & 81.56\\
    Chameleon~\cite{Chameleon_Team_Chameleon_Mixed-Modal_Early-Fusion_2024} & 11.84 & 2.07 & 8.32 & 10.08 $\pm$ 1.34 & 101.72 \\
    pipeline & 14.21 & 2.95 & 12.32 &  23.11 $\pm$ 1.85  & 61.38 \\
    end-to-end & \textbf{15.87} & \textbf{4.02} & \textbf{13.80} &  \textbf{25.61} $\pm$ \textbf{1.66}  & \textbf{59.40} \\

    \bottomrule
    \end{tabular}
    \label{tab:main_results}
\end{table}

\tablename~\ref{tab:main_results} reports the main results of different models.
First, we examine the performances of the pipeline and end-to-end model.
For the PhotoChat dataset, our end-to-end model can bring significant improvements for photo-sharing multi-modal dialogue, where gains are $12.14 - 10.68 = 1.46$, $2.98 - 2.10 = 0.88$ and $11.12 - 9.74 = 1.38$ on BLUE-1, BLUE-2 and Rouge-L for text response generation, and $14.59 - 13.71 = 0.88$ and $79.94 - 75.84 = 4.10$ on IS and FID for image response generation.
For the DialogCC dataset, the end-to-end model also advances the text response generation scores by $1.66$, $1.07$ and $1.48$ on BLUE-1, BLUE-2 and Rouge-L, and the image response generation scores by $2.50$ and $1.98$ on IS and FID, respectively.
All enhancements verify the effectiveness of our proposed end-to-end model compared with the basic pipeline model.

Second, we make comparisons between our end-to-end model and Divter, which is also a representative pipeline model.
For the DialogCC dataset, our end-to-end model significantly outperforms Divter, which might be due to the strong capabilities of text generation by Llama-3 and image generation by stable-diffusion-3.
Concretely, the end-to-end model boosts the performances of text and image response generation significantly, leading to increases on BLUE-1 by $7.79$, BLUE-2 by $2.13$, Rouge-L by $7.09$, IS by $16.42$ and FID by $49.10$, respectively. 
For the PhotoChat dataset, our end-to-end model still achieves better performances compared with naive Divter.
However, the best results of IS and FID are $15.8$ and $29.16$, which is from Divter$^\dagger$.
An important reason could be that \cite{sun-etal-2022-multimodal} employs extra image caption resources to train the model.
Sampling data from the two datasets might significantly influence the scores of IS and FID, because they are computed by the neural network trained on datasets with similar data distribution.

Third, we observe the results of other end-to-end multi-modal models.
The performance of GPT-4 is limited, because it is a close-source model and could not be fine-tuned on specific datasets.
We find its text responses are usually too long, but the ground-truths are oral expressions and short. 
For Chameleon, image generation results underperform our end-to-end model.
The reason could be that Chameleon opens only the text transformer parameter weights, but masks image generation parameter weights.

\subsection{Analyses}

\subsubsection{Ablation Experiment.}

\begin{table}[t]
    \centering
    \caption{Ablation experiment. }
    \setlength{\tabcolsep}{5pt}
    \begin{tabular}{c|ccc|cc}
    \toprule
    \multirow{2}{*}{\textbf{Model}} & \multicolumn{3}{c|}{\textbf{Text Generation}} & \multicolumn{2}{c}{\textbf{Image Generation}} \\
    & \textbf{BL-1} & \textbf{BL-2} & \textbf{Ro-L} & \textbf{IS $\uparrow$} & \textbf{FID $\downarrow$} \\
    \midrule

    end-to-end & \textbf{15.87} & \textbf{4.02} & \textbf{13.80} & \quad \textbf{25.61} $\pm$ \textbf{1.66} \quad & \textbf{59.40} \\
    \midrule
    -Image Perceptron & -1.36 & -0.78 & -1.18 & -1.29 & +0.70 \\
    -Image Generator & -0.37 & -0.21 & -0.42 & -1.94 & +1.47 \\

    \bottomrule
    \end{tabular}
    \label{tab:ablation_results}
\end{table}

The core of our end-to-end model is to replace the discrete image text descriptions in the pipeline model with dense vector representations, therefore gradient can propagate throughout the whole model without interruption.
We conduct an ablation experiment to show the influence of the proposed end-to-end techniques.
\tablename~\ref{tab:ablation_results} reports ablation experiment results.
-Image Perceptron means the end-to-end model ablates dense visual representations from the image perceptron and feeds discrete image text captions into the text dialogue generator.
-Image Generator means the text dialogue generator outputs discrete image text description for the diffusion model, which ablates the dynamic vocabulary transformation matrix and straight-through gumbel-softmax.
The two ablations significantly decrease the text and image response generation results, verifying the effectiveness of our proposed end-to-end model.

\subsubsection{Response Generation Performances for Different Speakers.}
\begin{figure}[t]
\centering
\begin{tikzpicture}
\definecolor{c1}{RGB}{162, 0, 037}
\definecolor{c2}{RGB}{0, 87, 0}
\definecolor{c3}{RGB}{0, 110, 175}
\definecolor{c4}{RGB}{180, 101, 4}

\begin{groupplot}[
    group style = {group size=5 by 1, horizontal sep=4pt, vertical sep=0pt},
    width=\columnwidth,
    height=4.0cm,
]

\nextgroupplot[
    ybar,
    xmin=0.3,
    xmax=3.7,
    ymin=12.8,
    ymax=17.5,
    width=3.2cm,
    height=4.0cm,
    bar width=0.3cm,
    title={BLUE-1},
    title style={font=\scriptsize, yshift=-0.31cm},
    ytick={0},
    yticklabels={0},
    yticklabel style={font=\scriptsize, xshift=0.05cm},
    x label style={font=\scriptsize, yshift=-0.1cm},
    xtick={1.1,2.9},
    xticklabels={pp, e2e},
    xticklabel style={font=\scriptsize, yshift=0.15cm},
    nodes near coords,
    nodes near coords style={font=\tiny, yshift=-0.1cm}
]

    \addplot[c1, fill=c1, bar shift=0pt, pattern=dots, pattern color=c1] coordinates {(0.9, 13.27)};
    \addplot[c2, fill=c2, bar shift=0pt, pattern=grid, pattern color=c2] coordinates {(2.5, 14.73)};
    \addplot[c3, fill=c3, bar shift=0pt, pattern=north west lines, pattern color=c3] coordinates {(1.5, 15.40)};
    \addplot[c4, fill=c4, bar shift=0pt, pattern=north east lines, pattern color=c4] coordinates {(3.1, 17.01)};

\nextgroupplot[
    ybar,
    xmin=0.3,
    xmax=3.7,
    ymin=2.5,
    ymax=4.4,
    width=3.2cm,
    height=4.0cm,
    bar width=0.3cm,
    title={BLUE-2},
    title style={font=\scriptsize, yshift=-0.31cm},
    ytick={0},
    yticklabels={0},
    yticklabel style={font=\scriptsize, xshift=0.05cm},
    x label style={font=\scriptsize, yshift=-0.1cm},
    xtick={1.1,2.9},
    xticklabels={pp, e2e},
    xticklabel style={font=\scriptsize, yshift=0.15cm},
    nodes near coords,
    nodes near coords style={font=\tiny, yshift=-0.1cm}
]
    \addplot[c1, fill=c1, bar shift=0pt, pattern=dots, pattern color=c1] coordinates {(0.9, 2.86)};
    \addplot[c2, fill=c2, bar shift=0pt, pattern=grid, pattern color=c2] coordinates {(2.5, 3.81)};
    \addplot[c3, fill=c3, bar shift=0pt, pattern=north west lines, pattern color=c3] coordinates {(1.5, 3.14)};
    \addplot[c4, fill=c4, bar shift=0pt, pattern=north east lines, pattern color=c4] coordinates {(3.1, 4.21)};

\nextgroupplot[
    ybar,
    xmin=0.3,
    xmax=3.7,
    ymin=11.3,
    ymax=15.5,
    width=3.2cm,
    height=4.0cm,
    bar width=0.3cm,
    title={Rouge-L},
    title style={font=\scriptsize, yshift=-0.31cm},
    ytick={0},
    yticklabels={0},
    yticklabel style={font=\scriptsize, xshift=0.05cm},
    x label style={font=\scriptsize, yshift=-0.1cm},
    xtick={1.1,2.9},
    xticklabels={pp, e2e},
    xticklabel style={font=\scriptsize, yshift=0.15cm},
    nodes near coords,
    nodes near coords style={font=\tiny, yshift=-0.1cm},
    legend style={at={(0.5, 1.33)}, anchor=north, legend columns=4, font=\scriptsize},
]

    \addplot[c1, fill=c1, bar shift=0pt, pattern=dots, pattern color=c1] coordinates {(0.9, 11.89)};
    \addplot[c2, fill=c2, bar shift=0pt, pattern=grid, pattern color=c2] coordinates {(2.5, 13.07)};
    \addplot[c3, fill=c3, bar shift=0pt, pattern=north west lines, pattern color=c3] coordinates {(1.5, 12.70)};
    \addplot[c4, fill=c4, bar shift=0pt, pattern=north east lines, pattern color=c4] coordinates {(3.1, 15.12)};

    \legend{pp speaker A, e2e speaker A, pp speaker B, e2e speaker B};

\nextgroupplot[
    ybar,
    xmin=0.3,
    xmax=3.7,
    ymin=21.1,
    ymax=28.8,
    width=3.2cm,
    height=4.0cm,
    bar width=0.3cm,
    title={IS},
    title style={font=\scriptsize, yshift=-0.31cm},
    ytick={0},
    yticklabels={0},
    yticklabel style={font=\scriptsize, xshift=0.05cm},
    x label style={font=\scriptsize, yshift=-0.1cm},
    xtick={1.1,2.9},
    xticklabels={pp, e2e},
    xticklabel style={font=\scriptsize, yshift=0.15cm},
    nodes near coords,
    nodes near coords style={font=\tiny, yshift=-0.1cm}
]
    \addplot[c1, fill=c1, bar shift=0pt, pattern=dots, pattern color=c1] coordinates {(0.9, 21.95)};
    \addplot[c2, fill=c2, bar shift=0pt, pattern=grid, pattern color=c2] coordinates {(2.5, 24.23)};
    \addplot[c3, fill=c3, bar shift=0pt, pattern=north west lines, pattern color=c3] coordinates {(1.5, 23.66)};
    \addplot[c4, fill=c4, bar shift=0pt, pattern=north east lines, pattern color=c4] coordinates {(3.1, 28.19)};

\nextgroupplot[
    ybar,
    xmin=0.3,
    xmax=3.7,
    ymin=54.1,
    ymax=66.2,
    width=3.2cm,
    height=4.0cm,
    bar width=0.3cm,
    title={FID},
    title style={font=\scriptsize, yshift=-0.31cm},
    ytick={0},
    yticklabels={0},
    yticklabel style={font=\scriptsize, xshift=0.05cm},
    x label style={font=\scriptsize, yshift=-0.1cm},
    xtick={1.1,2.9},
    xticklabels={pp, e2e},
    xticklabel style={font=\scriptsize, yshift=0.15cm},
    nodes near coords,
    nodes near coords style={font=\tiny, yshift=-0.1cm}
]
    \addplot[c1, fill=c1, bar shift=0pt, pattern=dots, pattern color=c1] coordinates {(0.9, 65.18)};
    \addplot[c2, fill=c2, bar shift=0pt, pattern=grid, pattern color=c2] coordinates {(2.5, 63.66)};
    \addplot[c3, fill=c3, bar shift=0pt, pattern=north west lines, pattern color=c3] coordinates {(1.5, 57.39)};
    \addplot[c4, fill=c4, bar shift=0pt, pattern=north east lines, pattern color=c4] coordinates {(3.1, 55.02)};

\end{groupplot}

\end{tikzpicture}
\caption{Response generation performances for different speakers.}
\label{fig:speaker_perf}
\end{figure}

There are two speakers in our photo-sharing multi-modal dialogue.
We examine the text and image response generation performances of the pipeline (pp) and end-to-end (e2e) model for the two speakers respectively.
\figurename~\ref{fig:speaker_perf} illustrates the experimental results.
Regardless of the pipeline model and end-to-end model, response performances of speaker B are superior to speaker A.
The reason might be from the instruction tuning stage of large language models, where the loss function is computed on the user side.
In our experiments, we employ the chat mode of large language models, and take the assistant side as speaker A, while the user side as speaker B.
Additionally, \figurename~\ref{fig:speaker_perf} also demonstrates that the end-to-end model advances all results compared with the pipeline model, verifying the effectiveness of our end-to-end model.

\subsubsection{Image Response Generation Performance for Different Gumbel-Softmax Temperature.}
\begin{figure}[t]
\centering
\begin{tikzpicture}

\definecolor{c1}{RGB}{162, 0, 037}
\definecolor{c2}{RGB}{0, 87, 0}
\definecolor{c3}{RGB}{0, 110, 175}
\definecolor{c4}{RGB}{180, 101, 4}

\begin{axis}[
    ymax=26.1,
    ymin=21.8,
    xmax=10.5,
    xmin=-0.5,
    ylabel={IS},
    y label style={font=\footnotesize, yshift=-0.6cm},
    ytick={22, 24, 26},
    yticklabels={22, 24, 26},
    y tick label style = {font=\footnotesize, xshift=0.1cm,},
    xlabel={Temperature of Gumbel-Softmax},
    x label style={font=\footnotesize, yshift=0.1cm},
    xtick = {0,2,4,6,8,10},
    xticklabels={$e^{-6}$, $e^{-5}$, $e^{-4}$, $e^{-3}$, $e^{-2}$, 1},
    x tick label style = {font=\footnotesize, yshift=0.0cm},
    width=8.2cm,
    height=4.2cm,
    legend style={at={(0.4, 0.6)}, anchor=north, legend columns=2, font=\footnotesize},
    axis y line*=left,
]

    \addplot[thick, solid, c1, mark=square*, mark size = 1.4pt, mark options={fill=c1, solid}] coordinates{
        (0, 23.51) (2, 24.86) (4, 25.56) (6, 24.72)  (8, 23.71) (10, 22.37)
    };\addlegendentry{IS}
    \addplot[thick, solid, c3, mark=o, mark size = 2pt, mark options={fill=c3, solid}] coordinates{
        (0, 62.51) (2, 60.86) (4, 59.63) (6, 60.72)  (8, 62.71) (10, 64.37)
    };\addlegendentry{FID}



\end{axis}

\begin{axis}[
    ymax=65,
    ymin=59,
    xmax=10.5,
    xmin=-0.5,
    ylabel={FID},
    y label style={font=\footnotesize, yshift=-8.5cm, rotate=180},
    ytick={60, 62, 64},
    yticklabels={60, 62, 64},
    y tick label style = {font=\footnotesize, xshift=-0.1cm,},
    x label style={font=\footnotesize, yshift=0.1cm},
    xtick = {0,2,4,6,8,10},
    xticklabels={$e^{-6}$, $e^{-5}$, $e^{-4}$, $e^{-3}$, $e^{-2}$, 1},
    x tick label style = {font=\footnotesize, yshift=0.0cm},
    width=8.2cm,
    height=4.2cm,
    axis y line*=right,
]

    \addplot[thick, solid, c3, mark=o, mark size = 2pt, mark options={fill=c3, solid}] coordinates{
        (0, 62.51) (2, 60.86) (4, 59.63) (6, 60.72)  (8, 62.71) (10, 64.37)
    };



\end{axis}

\end{tikzpicture}
\caption{Image response generation performance for different temperatures of gumbel-softmax.}
\label{fig:gs_temperature}
\end{figure}

The temperature of gumbel-softmax plays an important role in our end-to-end photo-sharing multi-modal dialogue generation model.
We conduct an analysis experiment to evaluate the relation between the temperature and image generation performance.
The results are demonstrated in \figurename~\ref{fig:gs_temperature}, where the x-axis denotes the temperature of gumbel-softmax and the y-axis denotes IS and FIS on the left and right, respectively.
When the temperature is 1, gumbel-softmax is equivalent to standard softmax.
As the temperature reduces lower from 1, the IS scores increase significantly, which verifies the effectiveness of gumbel-softmax for our end-to-end model.
The performance stops increasing after $e^{-4}$, which could be due to the weakened randomness.
There is a similar tendency for FID, but it is important to note that lower values are better for FID.

\section{Conclusion}
In this paper, we proposed the first end-to-end photo-sharing multi-modal dialogue model, which consists of three modules: text dialogue generator, image perceptron and image generator.  
We exploited an LLM as the backbone of the text dialogue generator.
For the image perceptron, Llama-Vision equipped the LLM with the ability to see images in the input end.
For the image generator, we applied the stable diffusion model to enable the LLM to draw images in the output end.
Particularly, we proposed the dynamic vocabulary transformation matrix and the straight-through gumbel-softmax technique to align the LLM and stable diffusion model and achieve end-to-end gradient propagation.
Finally, we conducted experiments on PhotoChat and DialogCC datasets to verify the effectiveness of our end-to-end model.
Compared with pipeline models, the end-to-end model gained state-of-the-art performances for photo-sharing multi-modal dialogue generation.

\bibliographystyle{IEEEbib}
\bibliography{icme2025references}

\end{document}